\documentclass{article} 
\usepackage{iclr2023_conference,times}

\usepackage{amsmath,amsfonts,bm}

\def\eqref#1{equation~\ref{#1}}

\def\1{\bm{1}}

\DeclareMathAlphabet{\mathsfit}{\encodingdefault}{\sfdefault}{m}{sl}
\SetMathAlphabet{\mathsfit}{bold}{\encodingdefault}{\sfdefault}{bx}{n}

\DeclareMathOperator*{\argmax}{arg\,max}
\DeclareMathOperator*{\argmin}{arg\,min}

\usepackage{hyperref}
\usepackage{url}
\usepackage{amsmath}
\usepackage{graphicx}
\usepackage{subcaption}
\usepackage{floatrow}
\usepackage{booktabs}

\title{Autoregressive Generative Modeling with Noise Conditional Maximum Likelihood Estimation}

\author{Henry Li \\
Yale University\\
\texttt{henry.li@yale.edu} \\
\And
Yuval Kluger \\
Yale University\\
\texttt{yuval.kluger@yale.edu} 
}

\begin{document}

\maketitle

\begin{abstract}
  We introduce a simple modification to the standard maximum likelihood estimation (MLE) framework. Rather than maximizing a single unconditional likelihood of the data under the model, we maximize a family of \textit{noise conditional} likelihoods consisting of the data perturbed by a continuum of noise levels. We find that models trained this way are more robust to noise, obtain higher test likelihoods, and generate higher quality images. They can also be sampled from via a novel score-based sampling scheme which combats the classical \textit{covariate shift} problem that occurs during sample generation in autoregressive models. Applying this augmentation to autoregressive image models, we obtain 3.32 bits per dimension on the ImageNet 64x64 dataset, and substantially improve the quality of generated samples in terms of the Frechet Inception distance (FID) --- from 37.50 to 12.09 on the CIFAR-10 dataset.
\end{abstract}

\section{Introduction}

\begin{figure}
\centering
\begin{subfigure}{.5\textwidth}
  \centering
  \includegraphics[width=.95\linewidth]{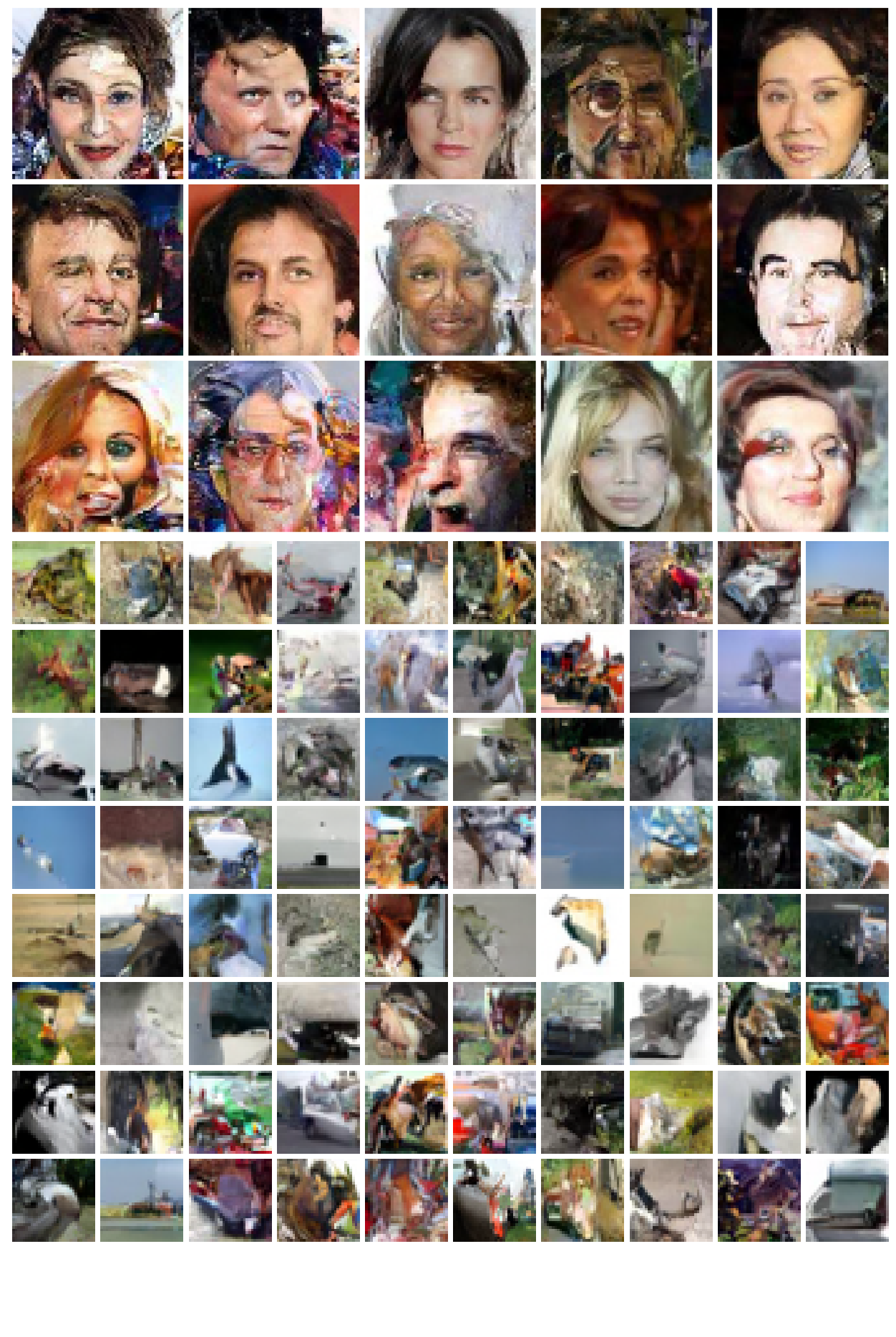}
\end{subfigure}%
\begin{subfigure}{.5\textwidth}
  \centering
  \includegraphics[width=.95\linewidth]{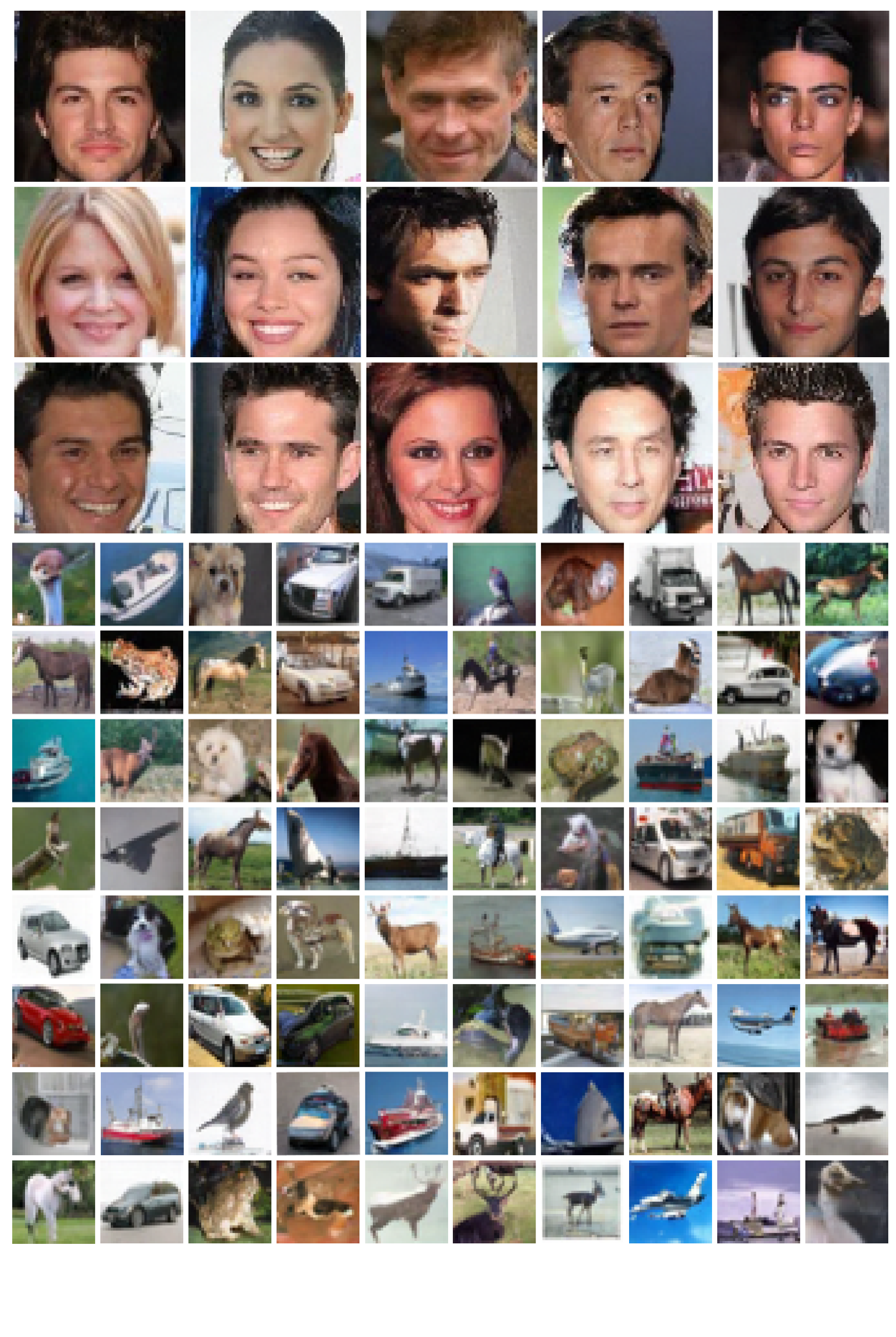}
\end{subfigure}
\vspace{-2em}
\caption{Generated samples on CelebA 64x64 \textbf{(above)} and CIFAR-10 \textbf{(below)}. Autoregressive models trained via vanilla maximum likelihood \textbf{(left)} are brittle to sampling errors and can quickly diverge, producing nonsensical results. Those trained by our  algorithm \textbf{(right)} are more robust and ultimately generate more coherent sequences.
}
\label{fig:main}
\end{figure}

Likelihood maximization models, \textit{i.e.}, models trained by maximizing log-likelihood, are a leading class of modern generative models. Of these, autoregressive models boast state-of-the-art performance in many domains, including images \cite{child2019generating}, text \cite{vaswani2017attention}, and audio \cite{oord2016wavenet}. 

However, while log-likelihood is broadly agreed upon as one of the most rigorous metrics for goodness-of-fit in statistical and generative modeling, models with high likelihoods do not necessarily produce samples of high visual quality. This phenomenon has been discussed at length by \cite{theis2015note, huszar2015not}, and corroborated in empirical studies \cite{grover2018flow, kim2022soft}.

Autoregressive models suffer an additional affliction: they have notoriously unstable dynamics during sample generation \cite{bengio2015scheduled, lamb2016professor} due to their sequential sampling algorithm, which can cause errors to compound across time steps. Such errors cannot usually be corrected \textit{ex post facto} due to the autoregressive structure of the model, and can substantially affect downstream steps as we find that model likelihoods are highly sensitive to even the most minor of perturbations.

Score-based diffusion models \cite{song2020score,ho2020denoising} offer a different perspective on the data generation process. Even though sampling is also sequential, diffusion models are more robust to perturbations because, in essence, they are trained as denoising functions \cite{ho2020denoising}. Moreover, the update direction in each step is unconstrained (unlike token-wise autoregressive models, which can only update one token at a time, and only once), meaning the model can correct errors from previous steps. However, diffusion models are poor likelihood models, as they cannot be trained via maximum likelihood\footnote{At best, they can optimize the lower bound of the likelihood as variational inference models \cite{song2021maximum,kingma2021variational}.}, and density evaluations are slow and have no closed form, requiring ODE/SDE solvers involving hundreds to thousands of calls to the underlying network. Thus we wonder: is there a conceptual middle ground?

In this paper, we offer such a framework. We further analyze the likelihood-sample quality mismatch in autoregressive models, and propose techniques inspired by diffusion models to alleviate it. In particular, we leverage the fact that the score function is naturally learned as a byproduct of maximum likelihood estimation. This allows a novel two-part sampling strategy with noisy sampling and score-based refinement.

Our contributions are threefold. 1) We investigate the pitfalls of training and inference under the log-likelihood maximization scheme, particularly regarding sensitivity to noise corruptions. 2) We propose a simple sanity test for checking the robustness of likelihood models to minor perturbations, and find that many models fail this test. 3) We introduce a novel framework for the training and sampling of likelihood maximization models that significantly improves the noise-robustness and generated sample quality. As a result, we obtain a model that can generate samples at a quality approaching that of diffusion models, without losing the maximum likelihood framework and $\mathcal{O}(1)$ likelihood evaluation speed of likelihood maximization models. 

The paper is structured as follows. We will cover background and related work (Section \ref{sec:bg_rw}), further investigate shortcomings of vanilla maximum likelihood (Section \ref{sec:pitfalls}), present the noise-conditional maximum likelihood (NCML) framework (Section \ref{sec:ncml}), and finally evaluate the robustness and performance of models trained under this framework (Section \ref{sec:experiments}).

\section{Background and Related Work}
\label{sec:bg_rw}

Let our dataset $\mathcal{X}$ consist of i.i.d. samples drawn from an unknown target density $\mathbf{x} \sim p_{data}(\mathbf{x})$. The goal of likelihood-based generative modeling is to approximate $p_{data}$ via a parametric model $p_{\boldsymbol\theta}$, where samples $x \sim p_{\boldsymbol\theta}$ can be easily obtained. 

\subsection{Background}
\label{sec:background}

We first discuss fundamental techniques for estimating and sampling from $p_\theta$ in generative modeling.

\textbf{Maximum Likelihood Estimation (MLE)}
The standard MLE framework consists in solving
\begin{equation}
    \argmax_{p_{\boldsymbol\theta}} \mathbb{E}_{\mathbf{x} \sim p_{data}} \log p_{\boldsymbol\theta}(\mathbf{x}) \approx \argmax_{p_{\boldsymbol\theta}} \frac{1}{|\mathcal{X}|} \sum_{\mathbf{x} \in \mathcal{X}} \log p_{\boldsymbol\theta}(\mathbf{x}).
    \label{eq:mle}
\end{equation}
When $p_{\boldsymbol\theta}$ is autoregressive, the likelihood of each sample can be further decomposed by the probabilistic chain rule, \textit{i.e.}, $ p_{\boldsymbol\theta}(\mathbf{x}) = p_{\boldsymbol\theta}(\mathbf{x}_1) p_{\boldsymbol\theta}(\mathbf{x}_2 | \mathbf{x}_i^1) \dots p_{\boldsymbol\theta}(\mathbf{x}_d | \mathbf{x}_{d-1} \dots \mathbf{x}_1)$
where $\mathbf{x}_k$ denotes the $k$th dimension of $\mathbf{x}$. 

Likelihood models draw samples $\mathbf{x} \sim p_{\boldsymbol\theta}$ one of two ways. Normalizing flows \cite{dinh2014nice,rezende2015variational} apply a series of invertible transformations to a latent variable $\mathbf{z} \in \mathbb{R}^d \sim p_{prior}$. On the other hand, autoregressive models sample $\mathbf{x}$ sequentially and coordinate-wise by drawing from each conditional likelihood.

\textbf{Score-based Modeling}
An alternative to MLE is score matching \cite{hyvarinen2005estimation}, \textit{i.e.}
\begin{equation}
    \argmin_{p_{\boldsymbol\theta}} \mathbb{E}_{x \sim p_{data}} ||\nabla_\mathbf{x} p_{\boldsymbol\theta}(\mathbf{x}) - \nabla_\mathbf{x} p_{data}(\mathbf{x})||^2,
    \label{eq:score_matching}
\end{equation}
where $\nabla_\mathbf{x} p(\mathbf{x})$ is also known as the \textit{(Stein) score} function.
Most score-based models sidestep the estimation $p_{\boldsymbol\theta}$ and directly estimate the score.
Sampling from $p_\theta$ can then be achieved via annealed Langevin dynamics \cite{song2019generative}, variational denoising \cite{ho2020denoising}, or reversing a diffusion process \cite{song2020score}. 

Of these, diffusion models \cite{song2020score} provide the most general framework for score-based sampling, as well as a means to recover $p_\theta$ by solving an ordinary differential equation (ODE). Here, each data point is modeled as a diffusion process, and can be seen as a function $\mathbf{x}: [0, T] \rightarrow \mathbb{R}^d$ such that $\mathbf{x}(0) \sim p_{data}$ and $\mathbf{x}(T) \sim p_{prior}$. The forward diffusion process is an Ito stochastic differential equation (SDE)
\begin{equation}
    d\mathbf{x} = \mathbf{f}(\mathbf{x}, t) + g(t) d\mathbf{w},
\end{equation}
for some drift and diffusion terms $\mathbf{f}$ and $g$, where $\mathbf{w}$ is the standard Wiener process. By a central result in \cite{anderson1982reverse}, this diffusion can be tractably reversed, and is moreover also a diffusion process whose SDE
\begin{equation}
    d\mathbf{x} = [\mathbf{f}(\mathbf{x}, t) + g^2(t) + \nabla_\mathbf{x} p_{t}(\mathbf{x})]dt + g(t) d\bar{\mathbf{w}},
    \label{eq:reverse_sde}
\end{equation}
depends again on $\mathbf{f}, g$, and additionally the noise-conditional score function, where $\bar{\mathbf{w}}$ is a backwards Wiener process. Sampling then consists of drawing $\mathbf{x}(T) \sim p_{prior}$, and solving the reverse diffusion process.

\textbf{A Hybrid Framework?} MLE and score-based models each have advantages and drawbacks when measured against the other. Score-based models exhibit unrivaled generative capabilities, producing images of state-of-the-art sample quality that beat out even GANs \cite{dhariwal2021diffusion}. However, they cannot be trained by maximum likelihood, and therefore do not have the same asymptotic guarantees as MLE (\textit{e.g.}, efficiency, functional equivariance, sufficiency). Moreover, likelihoods have no closed-form expression, thus requiring approximate ODE/SDE solvers and hundreds to thousands of function evaluations to evaluate.

MLE models, on the other hand, are best-in-class in terms of model log-likelihood and log-likelihood computation speed --- which, outside of its inherent value, has proven to be useful in many downstream tasks, including anomaly and out-of-distribution (OOD) detection \cite{ren2019likelihood}, adversarial defense \cite{song2017pixeldefend}, among others. Additionally, they enjoy the guarantees that come with maximum likelihood estimation. However, they are well-known have subpar generative sample quality, which we further investigate in Section \ref{sec:pitfalls}.


These shortcomings raise the natural question of whether there exist frameworks that bridge the gap between the methods. In this paper, we explore this question.


\subsection{Related Work}
\label{sec:related_work}

A number of other works combine score- and energy-based modeling with autoregressive architectures. \cite{hoogeboom2021autoregressive} propose an order-agnostic autoregressive model for simulating \textit{discrete} diffusions via a variational lower bound. \cite{meng2020autoregressive} use unnormalized autoregressive models to learn distributions in an augmented score-matching framework. \cite{nash2019autoregressive} design an energy-based model with an autoregressive structure such that the normalizing constant can be estimated coordinate-wise via importance sampling. However, each approach relinquishes the ability to compute exact likelihoods in their framework, which is one of the motivating advantages of using autoregressive structures.

A closely related vein of research explores alternative training and inference strategies so as to improve sampling stability in autoregressive models. \cite{bengio2015scheduled} propose training models with a mixture of true and generated samples over time, where the proportion of generated samples gradually grows to take up the majority of training sequences. This is initially promising, except \cite{huszar2015not} note that this technique is biased and not guaranteed to produce solutions that converge on the true distribution. \cite{lamb2016professor} subsequently suggest incorporating an adversarial loss provided by a discriminator that "teaches" the model to produce more realistic samples over multiple sampling steps. Both techniques again depart from the maximum likelihood framework. Perhaps most similar to our approach is \cite{jayaram2021parallel}, who also suggest sampling via the score function with Langevin dynamics, but they crucially do not train with noise, which \cite{song2019generative} found to be essential for stable sampling in a Langevin algorithm. As a result, they are not able to sample images unconditionally.

Ultimately, our approach uniquely provides a principled and generalized framework for modeling stochastic processes (including diffusions) that retains the asymptotic guarantees of maximum likelihood estimation while producing samples of superior quality.

\section{The Pitfalls of Maximum Likelihood}
\label{sec:pitfalls}
We first show that density models trained to maximize the standard log-likelihood are surprisingly sensitive to minor perturbations. We then discuss why this is bad for generative modeling performance.

\subsection{A Simple Sanity Test}
\label{sec:sanity}
Consider the class of minimally corrupted probability densities we call $p_\pi$, where
\begin{equation}
    p_\pi = p_{data} \ast p_{mult_{\{-1, 0, 1\}}(\pi/2, 1 - \pi, \pi/2)}, \hspace{.5in} \pi \in [0, 1].
    \label{eq:p_pi}
\end{equation}
Here, $\ast$ denotes the convolution operator, and $p_{mult_{\{a, b, c\}}(\alpha, \beta, \gamma)}$ is the density a $d$-dimensional multinomial distribution taking on $a$, $b$, and $c$ with probabilities $\alpha$, $\beta$, and $\gamma$, respectively. $p_\pi$ is \textit{minimally corrupted} in the sense that, if $p_{data}$ is an integer-discretized distribution (say, 8-bit images, or 16-bit digital audio signals), $p_\pi$ describes the distribution of points in $p_{data}$ that have had their least significant bit incremented or decremented with probability $\pi$.

\begin{figure}
\centering
\includegraphics[width=\linewidth]{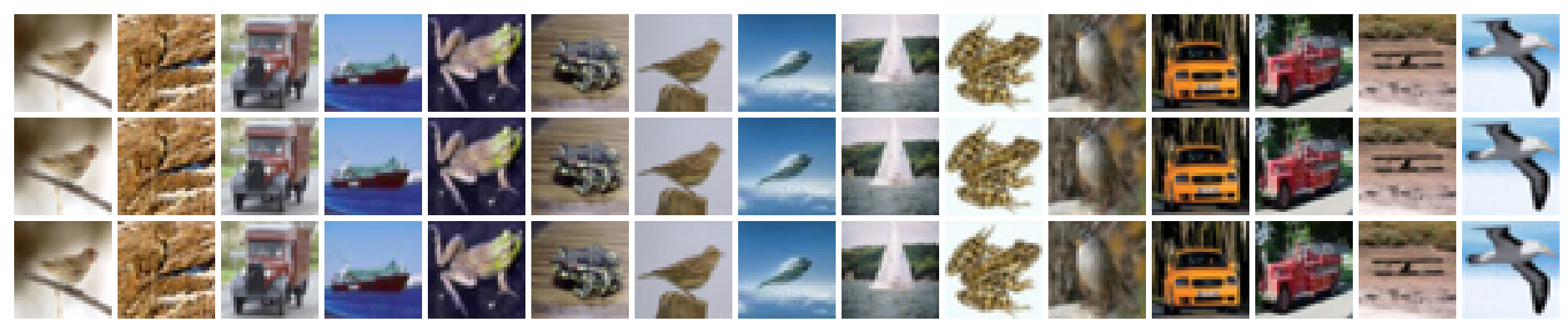}
\caption{Images from CIFAR-10 \textbf{(top)} versus their $p_\pi$ perturbed counterparts for $\pi = 0.5$ \textbf{(middle)} and $\pi = 1$ \textbf{(bottom)}. The differences are nearly indistinguishable to the human eye, yet cause drastic deviations in average log-likelihood for standard likelihood models (Section \ref{sec:pitfalls} and Table \ref{table:main}).}
\label{fig:p_pi}
\end{figure}

In 8-bit images, the difference between samples drawn from $p_\pi$ and $p_{data}$ is almost imperceptible to the human eye, even for $\pi=1$ (see Fig \ref{fig:p_pi}). However, for likelihood models, this perturbation drastically increases the negative log-likelihood of the data under the model (see Table \ref{table:main}), to the point that it significantly undermines (if not outright nullifies) any recent advances in density estimation. This basic inconsistency suggests that the learned density of many standard likelihood models is brittle and overly emphasizes bit-level statistics that have little influence on the inherent content of the image.

\subsection{Why We Should Care}
We provide three reasons for why failing this test is problematic, especially for autoregressive models.

First, noise is natural --- and being less robust to noise also means being a poorer fit to natural data, especially in the ways that matter to the end user. Outside of the log-likelihood, measures of generative success in generative models fall under two categories: qualitative assessments (\textit{e.g.}, the no-reference perceptual quality assessment \cite{wang2002no} or 'eyeballing' it) and quantitative heuristics (\textit{e.g.}, computing statistics of hidden activations of pretrained CNNs \cite{inception, frechet, sajjadi2018assessing}). Both strategies either rely directly on the human visual system, or are known to be closely related to it \cite{cnn_brain1,cnn_brain2,cnn_brain3,cnn_brain4,cnn_brain5}. Therefore, implicit in the use of these criteria is assuming the existence of a human (or human-like) model of images $q_{human}$, where $q_{human} \approx p_{data}$ \cite{huszar2015not}. This assumption, as well as the fact that we find samples from $p_\pi$ nearly indistinguishable from $p_{data}$, whereas $p_\theta$ finds them very different, suggests that $p_{data} \approx q_{human} \neq p_\theta$.

Second, it suggests that we may need to re-evaluate our current research trajectory in likelihood modeling on images. Perhaps more surprising than failing the test is the fact all likelihood models fail the test more or less equally. As touched upon in Section \ref{sec:sanity}, Table \ref{table:main} shows that recent gains in autoregressive likelihood modeling are largely diminished when evaluating the average model likelihood of $p_\pi$ for $\pi = 0.5$ and $\pi = 1$. For example, the difference in likelihood between one of the first autoregressive likelihood models evaluated on CIFAR-10 \cite{salimans2017pixelcnn++} and the current state of the art \cite{child2019generating} is 0.12 bits per dim (bpd). However, under our perturbation, the difference is 0.02 bpd, which is more than a 6x reduction. Similarly, the difference between the state-of-the-art and normalizing flow models, which are known to be inferior to autoregressive models in terms of likelihood, is also significantly reduced. This indicates that much of the progress in recent years leans heavily towards improvements in modeling the least significant bit, which ultimately bears little significance to the content of the image.

Third, generative sample quality suffers. This holds for general likelihood models, given what we argue in the first point --- namely $p_\theta$ is very different from $q_{human}$. However, noise-sensitivity is doubly problematic in autoregressive models. Due to the sequential nature of autoregressive sampling and the fact that models are presented only with data from the \textit{true} distribution during training, autoregressive models are already known to be poorly-equipped to handle the sequences they encounter during sampling \cite{bengio2015scheduled}. Any sampling error introduces a shift in distribution that can affect the model's predicted likelihood of downstream tokens. This will increase the risk of mis-sampling the next token, which, in turn, further affects the downstream likelihoods. This is related to the well-known \textit{covariate shift} phenomenon \cite{shimodaira2000improving}. Sensitivity to minor perturbations only exacerbates the problem, and Table \ref{table:main} shows that in vanilla autoregressive likelihood models, mis-sampling pixels by even a single bit can cause drastic changes to the overall likelihood. This can explain why such models commonly produce nonsensical results (Fig \ref{fig:main}).

For these reasons, we find that improving noise-robustness is of central importance to likelihood-based generative modeling, and especially likelihood-based autoregressive generative models.

\section{Noise Conditional Maximum Likelihood}

\label{sec:ncml}

\begin{figure}
\centering
\includegraphics[width=\linewidth]{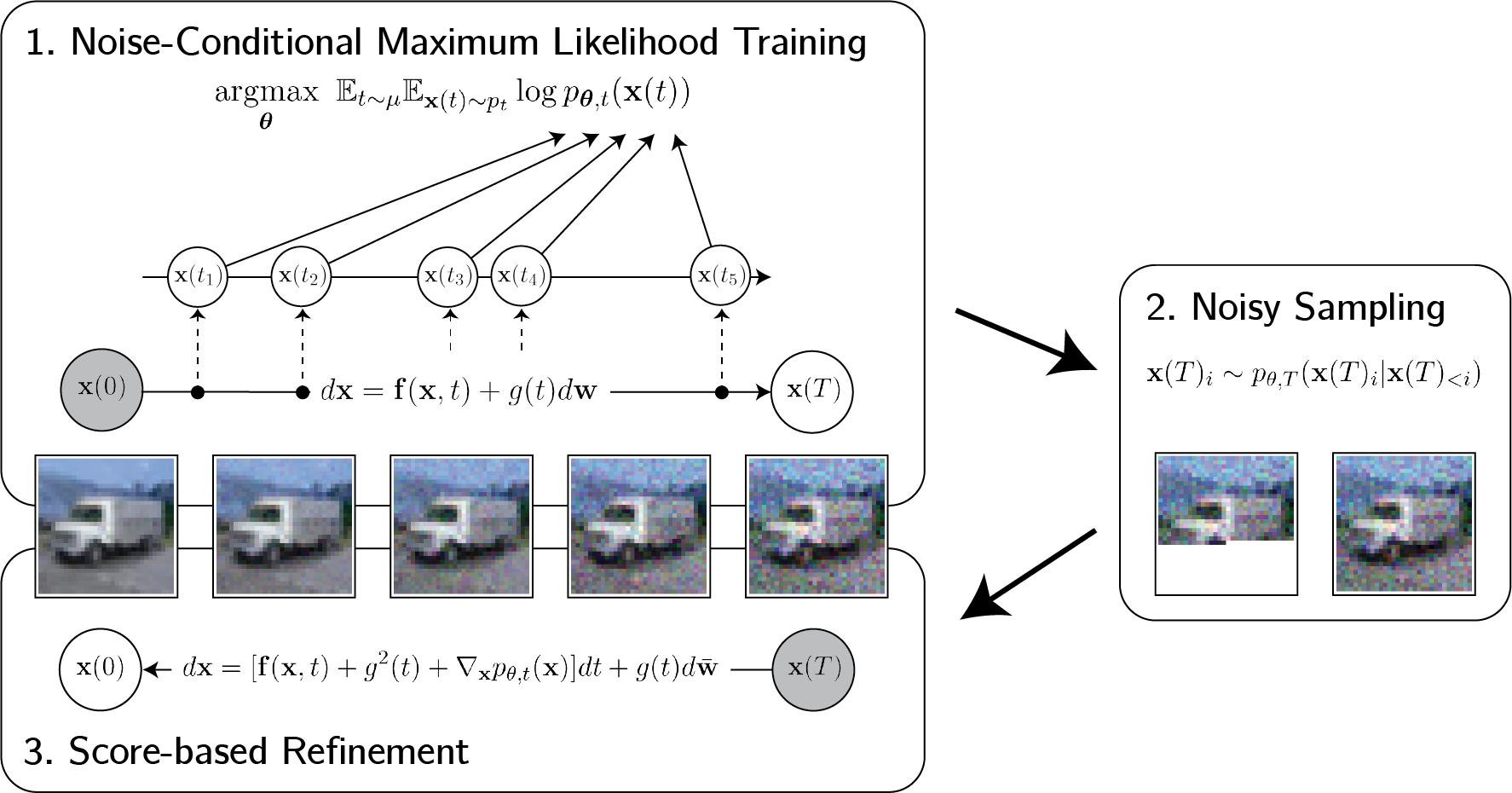}
\caption{An overview of the NCML generative algorithm. There are three steps: 1) We train the noise conditional density model $p_{\theta, t}$ via NCML. 2) We sample from the modeled noisy distribution, \textit{i.e.}, $p_{\theta, t}$ with some $t > 0$. 3) We refine the sample by solving a reverse diffusion involving the learned score function $\nabla_x p_{\theta, t}(x)$.}
\label{fig:cartoon}
\end{figure}

To alleviate the problems discussed in Section \ref{sec:pitfalls}, we propose a simple modification to the standard objective in maximum likelihood estimation. Rather than evaluating a single likelihood as in 
the vanilla formulation, we consider a family of noise conditional likelihoods 
\begin{equation}
    \mathbb{E}_{t \sim \mu} \mathbb{E}_{\mathbf{x} \sim p_{t}} \log p_{\boldsymbol\theta, t}(\mathbf{x}),
    \label{eq:ncmle}
\end{equation}
where $p_t$ is a stochastic process indexed by noise scales $t$ modeling a noise-corrupting process on $p_{data}$, and $\mu$ is a distribution over such scales. We call this the \textbf{noise conditional maximum likelihood (NCML)} framework. In general, Eq \ref{eq:ncmle} is an all-purpose plug-in objective that can be used with any likelihood model adapted to accept a noise conditioning vector\footnote{Though a continuous likelihood or a discretization of it (e.g. normalizing flows \cite{dinh2014nice} or autoregressive models with non-softmax probabilities \cite{salimans2017pixelcnn++,li2022neural}) is necessary for computation of the score function.}. We now explore various perspectives of NCML that may help with reasoning about this framework.

\textbf{NCML as Regularized Maximum Likelihood} When the set of noise scales $t \in \mathcal{T}$ is finitely large, a natural and mathematically equivalent formulation of NCML is as a form of data- and model- dependent regularization of the standard maximum likelihood estimation objective:
\begin{equation}
    \mathbb{E}_{t \sim \mu} \mathbb{E}_{\mathbf{x} \sim p_{t}} \log p_{\boldsymbol\theta, t}(\mathbf{x}) \propto  
    \underbrace{
        \mathbb{E}_{\mathbf{x} \sim p_{data}} \log p_{\boldsymbol\theta, 0}(\mathbf{x})
        \vphantom{\sum_{t \in \mathcal{T}}}
    }_{\text{MLE objective}}
    +
    \underbrace{
        \sum_{t \in \mathcal{T}} \lambda_t \mathbb{E}_{\mathbf{x} \sim p_{t}} \log p_{\boldsymbol\theta, t}(\mathbf{x})
    }_{\text{regularization term}},
\end{equation}
where $\mathcal{T}$ comprises the set of nonzero noise scales and $\lambda_t := \mu(t) / \mu(0)$. Clearly, the standard likelihood can be considered a special case of our proposed method where $\lambda_t = 0$ for all $t \in \mathcal{T}$. Furthermore, since the NCML framework can simply be seen as formulating $|\mathcal{T}|$ simultaneous and separate MLE problems, it retains all the statistical properties of standard MLE.

Of course, any form of regularization introduces bias to the model framework. Whereas L0/L1/L2 regularizations bias towards solutions of minimal or sparse weight norm, our experiments suggest that NCML biases towards solutions that are less sensitive to noisy perturbations (see Section \ref{sec:ncml_robustness}). 


\textbf{NCML as a Diffusion Model} Letting $t$ be the time index of a diffusion process, our approach becomes closely related to score-based diffusion models \cite{song2020score}, albeit with two crucial differences.

First, instead of merely estimating the noise-conditional score $\nabla_x \log p_t(x)$ for $t \in \mathcal{T}$, we directly estimate $p_t$ itself. However, $\nabla_x \log p_t(x)$ is still learned as a by-product of NCML, as it is the derivative of the log of the learned quantity. We may then access our approximated score via standard backpropagation techniques. Therefore, like diffusion models, we can draw samples via Langevin dynamics. This provides an alternative strategy for sampling from $p_{\theta, t}$, which we explore in \ref{sec:sampling}.

Second, we need not design our diffusion so that $p_T$ approximates the limiting stationary distribution of the process. This is necessary in diffusion models as the limiting prior is the only tractable distribution to initialize the sampling algorithm with. Since we have learned the density itself for all $t \in \mathcal{T}$, we may initialize from any point of the diffusion, which increases the flexibility of the sampling strategy, and can drastically reduce the steps required to solve the reverse diffusion.

For our models, we consider the three diffusion processes proposed in \cite{song2020score}: variance exploding (VE), variance preserving (VP), or sub-variance preserving (sub-VP), and choose $\mu$ to be the uniform distribution over $\mathcal{T}$. Due to space constraints, we refer to the aforementioned paper for more details on these SDEs.

\subsection{Noise-robustness of NCML Models}
\label{sec:ncml_robustness}

\begin{figure}
\centering
\begin{subfigure}{.33\textwidth}
  \centering
  \includegraphics[width=.95\linewidth]{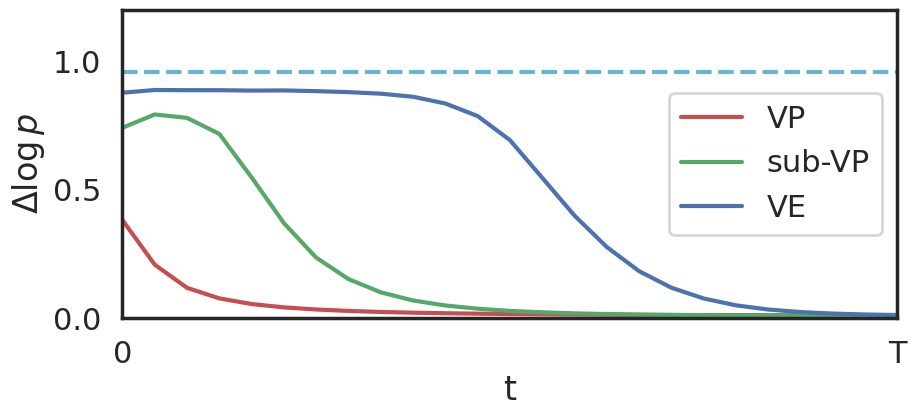}
\end{subfigure}%
\begin{subfigure}{.33\textwidth}
  \centering
  \includegraphics[width=.95\linewidth]{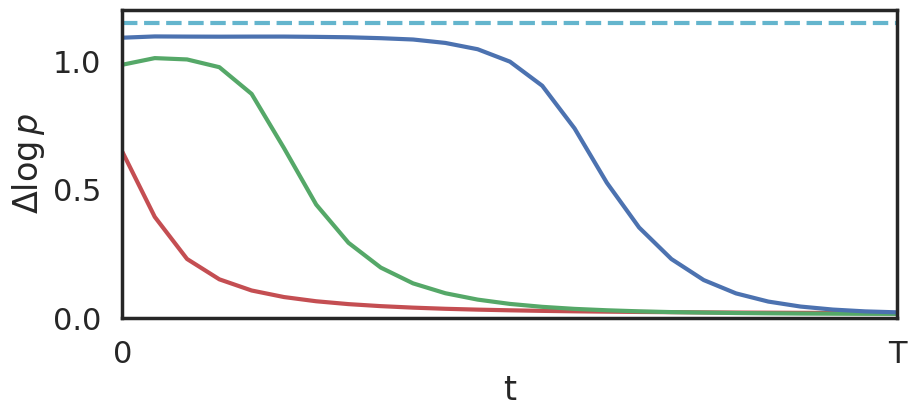}
\end{subfigure}
\begin{subfigure}{.33\textwidth}
  \centering
  \includegraphics[width=.95\linewidth]{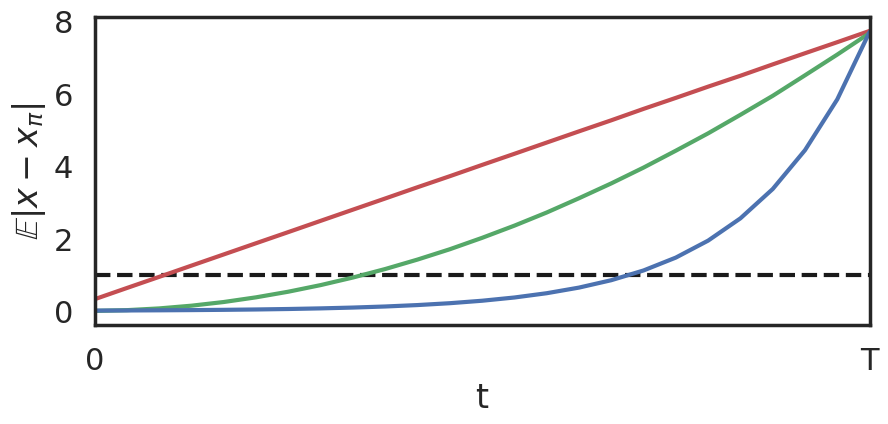}
\end{subfigure}
\caption{Noise robustness of NCML-trained models with Variance Preserving (VP), sub- Variance Preserving (sub-VP), and Variance Exploding (VE) noise schedules, measured in terms of $\Delta \log p$ (defined in \ref{eq:delta_logp}) between $p_{data}$ and $p_\pi$, for $\pi = 0.5$ \textbf{(left)} and $\pi = 1$ \textbf{(middle)}. \textbf{(Right)} shows that robustness is closely related to the average absolute perturbation per pixel of each noise schedule, as a function of $t$. More details in Section \ref{sec:ncml_robustness}.
}
\label{fig:robustness}
\end{figure}

We find that models trained via NCML are more robust to noise. Surprisingly, this is not only the case for noise conditions $t > 0$, where the model is exposed to noisy samples. Indeed, even in the $t = 0$ condition, noise-robustness is improved (Table \ref{table:main}), and our models surpass state-of-the-art likelihood models on minimally perturbed data (as defined in Eq \ref{eq:p_pi}) in terms of average log-likelihood. This is somewhat unexpected: by passing the noise condition, the model should theoretically be able to separate the NCML loss into distinct problems at each noise scale. If this occurs, then the $t = 0$ case should simplify to a vanilla MLE problem, where  behavior would not differ from standard likelihood models. Yet this is not the case.

We believe that this lends credence to the regularization perspective provided in \ref{sec:ncml}. In essence, noise robustness is explicitly enforced for $t > 0$. For $t = 0$, NCML leverages the limited capacity of the underlying network to implicitly impose robustness. To quantify the noise-robustness at each $t$, we define the simple measure $\Delta \log p$ as the absolute difference between the negative model log likelihood (as measured in bits per dimension) evaluated on $p_{data}$ minus that on $p_\pi$, \textit{i.e.},
\begin{equation}
    \Delta \log p := |\mathbb{E}_{p_{data}} \log p_{\boldsymbol\theta} - \mathbb{E}_{p_\pi} \log p_{\boldsymbol\theta}|.
    \label{eq:delta_logp}
\end{equation}
The left and middle graphs in Fig \ref{fig:robustness} show $\Delta \log p$ as a function of $t$ for $\pi = 0.5$ and $\pi = 1$, respectively. Here, we can clearly see that noise robustness increases with increasing $t$. Moreover, the regularization effect of NCML enforces greater robustness than competing models even at $t=0$, as seen by the dotted line showing the next lowest $\Delta \log p$, indicating that NCML enforces some degree of noise-robustness as regularization.

As one may expect, the correlation between noise-robustness and $t$ follows closely to the noise schedules of VP, sub-VP, and VE SDEs in $[0, T]$ respectively. This can be seen in the rightmost plot of Figure \ref{fig:robustness}, which shows the average absolute perturbation per pixel of each SDE over time. The dotted line represents $1$, \textit{i.e.}, the absolute perturbation of corrupted images in our sanity test $p_\pi$. The point at which each model attains robustness to $p_\pi$ corrupted noise is more or less the same time the noise schedule begins to perturb each pixel by at least one bit, on average.

The increased noise-robustness of NCML-trained models at larger $t$ motivates our improved autoregressive sampling algorithm, which we introduce below.

\subsection{Sampling with Autoregressive NCML Models}
\label{sec:sampling}

Our framework allows for two sampling strategies. The first is to draw directly from the noise-free distribution $p_{\boldsymbol\theta, 0}$, in which case the conditional likelihood simplifies to a standard (unconditional) likelihood, and sampling is identical to that for a vanilla autoregressive model.

However, as discussed in Section \ref{sec:pitfalls}, this strategy is unstable and tends to quickly accumulate errors. This motivates an alternative two-part sampling strategy, which involves drawing from $p_{\theta, t}$ for $t > 0$ (the \textit{noisy sampling} phase), then solving a reverse diffusion process back to $t = 0$ (the \textit{score-based refinement} phase). The tractability of the latter is due to the fact that NCML-trained models learn the score as a byproduct of likelihood estimation. This is identical to the sampling procedure in score-based diffusion models \cite{song2020score}, except for the key difference that we need not initialize with the prior distribution, as we can sample from any $t \in \mathcal{T}$.

This two-part strategy has several benefits. First, we saw in Section \ref{sec:ncml_robustness} that NCML-trained models are more robust to noise at higher $t$, which improves stability during sampling. This is reflected in the improved FID of NCML-trained NCPNs compared to ML- (\textit{i.e.}, maximum likelihood-) trained NCPNs in Table \ref{table:main}. Second, the refinement phase allows the model to make fine-tuned adjustments to the sample, which can further improve quality. This is not possible in standard autoregressive models by construction.



\section{Experiments}
\label{sec:experiments}

\begin{table}
  \caption{Results on CIFAR-10 and ImageNet 64x64. Negative log-likelihood (NLL) is in bits per dimension. Lower is better. *NLL with $\pi=0$ is equivalent to NLL of the original data.}
  \centering
  \label{table:main}
  \begin{tabular}{lccccccc}
    \toprule
                &\multicolumn{4}{c}{CIFAR-10}   &\multicolumn{3}{c}{ImageNet 64x64}                   \\
    \cmidrule(r){2-5}
    \cmidrule(r){6-8}
    Model   & FID   & NLL   & NLL  & NLL   & NLL   & NLL  & NLL \\
            &       & $\pi=0$*    & $\pi=0.5$     & $\pi=1$   & $\pi=0$*    & $\pi=0.5$     & $\pi=1$ \\
    \toprule
    \textbf{ELBO}                               \\
    \toprule
    VDM            & 7.41   & \textbf{2.49} & \textbf{3.75} & \textbf{3.97} & \textbf{3.40} & 3.76 & 3.88       \\
    ScoreFlow       & \textbf{5.40}   & 2.90 & 3.82 & 3.99  & - & - & -         \\
    VDVAE        & -              & 2.84 & 3.90 & 4.10 & 3.52 & \textbf{3.66} & \textbf{3.82}                                                \\
    \toprule 
    \textbf{Likelihood}                               \\
    \toprule
    Flow++     & -       & 3.09  & 3.86 & 4.08 & 3.69 & 3.82 & 3.99  \\
    DenseFlow & 48.15 & 2.98 & 3.80 & 4.02 & 3.35 & 3.68 & 3.85 \\
    PixelCNN++ & 55.72   & 2.92 & 3.84 & 4.01 & 3.52 & 3.84 & 4.00 \\
    PixelSNAIL     & 36.62       & 2.85 & 3.83  & 3.99  & - & - & -\\
    Sparse Transformer     & 37.50       & \textbf{2.80}  & 3.82 & 3.98 & 3.44 & 3.73 & 3.89 \\
    \toprule
    NCPN (ML)    & 46.72   & 2.91  & 3.83 & 3.99 & 3.49 & 3.68 & 3.88 \\
    NCPN (NCML-VE)    & 32.71   & 2.87  & 3.75 & 3.95 & \textbf{3.32}    & 3.67  & 3.85    \\
    NCPN (NCML-subVP) & 23.42   & 2.95  & 3.69 & 3.94  & 3.36    & 3.66  &  3.82  \\
    NCPN (NCML-VP)    & \textbf{12.09}   & 3.20  & \textbf{3.62} & \textbf{3.91} & 3.43     & \textbf{3.63} & \textbf{3.79} \\
    \bottomrule
  \end{tabular}
\end{table}

We demonstrate that incorporating noise in the maximum likelihood framework provides significant improvements in terms of both density estimation and sample generation. In all experiments, we fix $p_t$ to be one of the variance exploding (VE), variance preserving (VP), or sub-variance preserving (sub-VP) SDEs, and $\mu$ to be the uniform distribution over $t \in \mathcal{T}$. For our architecture, we introduce the noise conditional pixel network (NCPN), which consists of a PixelCNN++ \cite{salimans2017pixelcnn++} backbone with added attention layers. 

\textbf{Unconditional Modeling}
We evaluate our models on minimally perturbed transformations (see Section \ref{sec:sanity}) of unconditional CIFAR-10 and ImageNet 64x64 for $\pi \in \{0, \frac{1}{2}, 1\}$, where we note that $p_{\pi} = p_{data}$ when $\pi = 0$. All noise conditional models, \textit{i.e.}, ours, VDM \cite{kingma2021variational}, and ScoreFlow \cite{song2021maximum}, are evaluated at $t=0$. We show our results in Table \ref{table:main}. We additionally evaluate our model on the CelebA 64x64 dataset; autoregressive     comparisons on this dataset are limited, so we defer results to the appendix.

On the standard unperturbed dataset $\pi = 0$, our models attain competitive likelihoods on CIFAR-10 and state-of-the-art likelihoods on ImageNet 64x64. On all perturbed datasets, our models achieve state-of-the-art likelihoods. Furthermore, we significantly improve on the state-of-the-art in terms of sample quality among likelihood maximization models on CIFAR-10, from 37.50 to 12.09 in terms of FID. In general, our results indicate that the NCML framework improves generative modeling performance of the underlying models in terms of both test log-likelihoods and sample quality.

\begin{figure}
\centering
\begin{subfigure}{.5\textwidth}
  \centering
  \includegraphics[width=.95\linewidth]{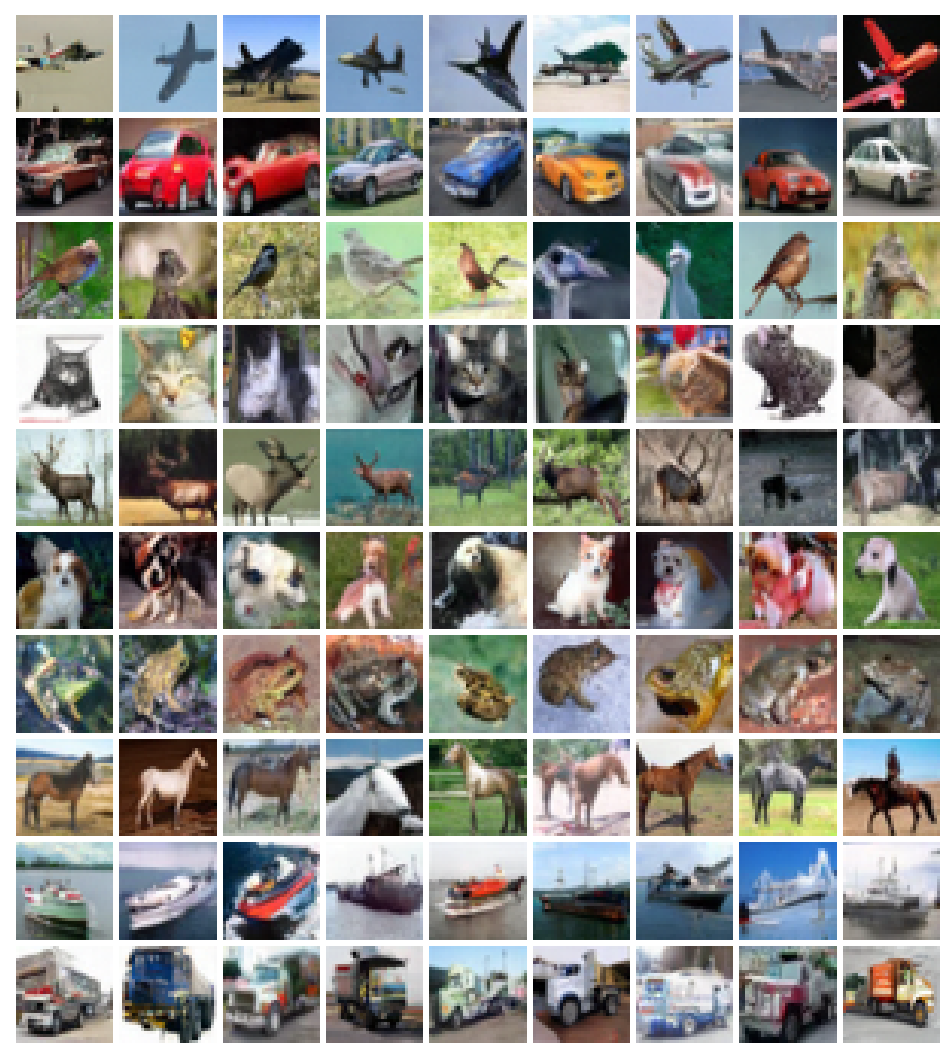}
\end{subfigure}%
\begin{subfigure}{.5\textwidth}
  \centering
  \includegraphics[width=.95\linewidth]{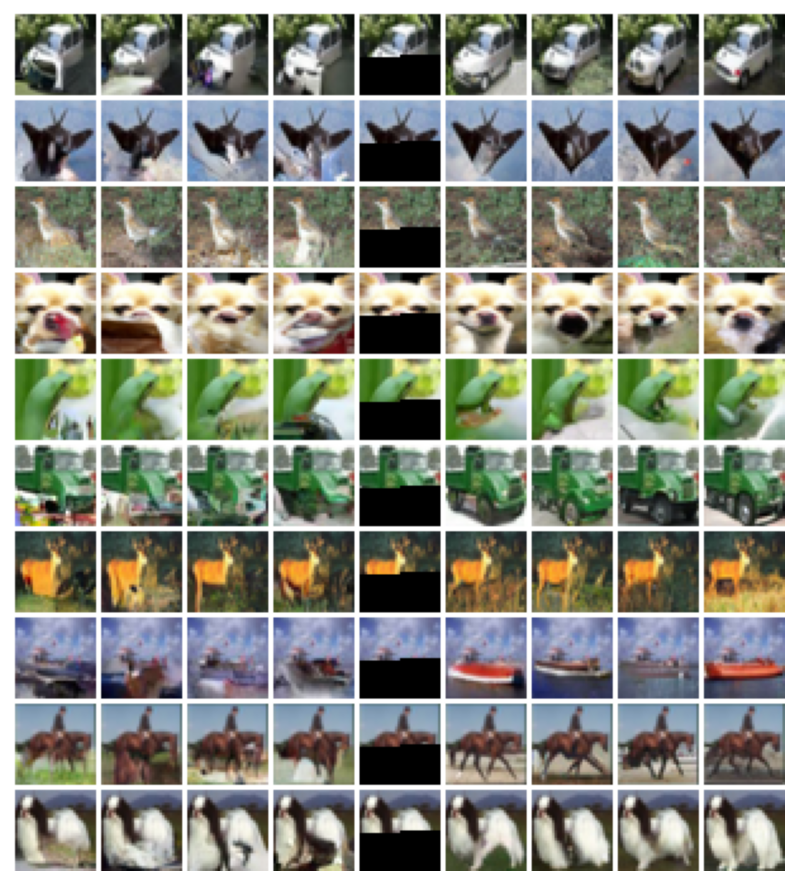}
\end{subfigure}
\caption{Class-conditional sampling on CIFAR-10 \textbf{(left)}. Image completion on CIFAR-10 \textbf{(right)}. See Section \ref{sec:experiments} for more details.
}
\label{fig:conditional_exp}
\end{figure}

\textbf{Class-conditional Modeling}
We find that NCML-trained models exhibit stable sampling even in class-conditional generation. For this experiment, we train a class-conditional model under our framework on the CIFAR-10 dataset. We show our results in Figure \ref{fig:conditional_exp}, where each row shows data sampled from a different class of CIFAR-10. 

\textbf{Image Completion} We further examine NCML-trained models in a controllable generation context through the image completion task, which involves conditioning an autoregressive model on the first half of an image, and drawing the second half from the modeled conditional distribution. We again use the CIFAR-10 dataset, and compare models trained under our framework against those trained with MLE. All images are taken from the test set to minimize data memorization. Our results are in figure \ref{fig:conditional_exp} (right). 
The leftmost and rightmost four columns are outputs generated by an MLE-trained model and an NCML-trained model, respectively, with the middle column depicting the masked input to the sampling algorithm.
Both models use the same architecture. Our model demonstrates improved stability over the course of sampling, and produces completed images with greater realism and fidelity, while maintaining a high diversity of sampled trajectories.

\section{Conclusion and Further Work}

We proposed a simple sanity test for checking the robustness of likelihoods to visually imperceptible levels of noise, and found that most models are highly sensitive to perturbations under this test. We argue that this is further evidence of a fundamental disconnect between likelihoods and other sample quality metrics. To alleviate this issue, we developed a novel framework for training likelihood models that combines autoregressive and diffusion models in a principled manner. Finally, we find that models trained under this setting have substantial improvements in both training and sampling.

While models trained under the NCML framework show greater invariance to imperceptible noise, they are by no means robust, indicating that the underlying model still differs significantly from the theoretical human model $q_{human}$ proposed in \cite{huszar2015not}. We hope that further research can help close this gap, and furnish us with a more intuitive grasp on the maximum likelihood as a framework for assessing goodness-of-fit in generative models.



\bibliography{main}
\bibliographystyle{iclr2023_conference}

\clearpage
\appendix

\section{Appendix}

\subsection{Additional Experimental Details}
For experiments on CIFAR-10 and ImageNet 64x64, we compare against \cite{kingma2021variational,song2021maximum,child2020very,ho2019flow++,grcic2021densely,salimans2017pixelcnn++,chen2018pixelsnail,child2019generating}. Some results could not be included due to the irreproducibility of the techniques. There is limited existing work on likelihood-based modeling on CelebA 64x64, so we do not provide comparisons here.

Our proposed NCPN architecture consists of the PixelCNN++ backbone \cite{salimans2017pixelcnn++} with axial attention layers \cite{ho2019axial} after each residual block. We retain the hyperparameters of PixelCNN++, changing only the dropout on the CIFAR-10 dataset (from 0.5 to 0.25), which we reduced due to the regularization properties of NCML. For the axial attention layers, we use 8 heads and skip connection rescaling as in \cite{song2020score}. Finally, we add noise conditioning to each residual block via a Gaussian Fourier Projection layer, much like \cite{ho2020denoising,song2020score}.

For our NCML-trained models, the diffusion times of the VE, VP, and sub-VP SDEs were chosen to be $T=0.5$, $T=0.1$, and $T=0.025$, respectively. The values are somewhat arbitrary, but selected such that the standard deviation of the per-pixel differences between samples in $p_{data}$ and their noised counterparts in $p_T$ was $\approx 10$. We suspect that further improvements can be made to the empirical results if these numbers were chosen more judiciously.

All NCPN models were trained on RTX 2080 Ti GPUs for 500,000 iterations. This is approximately 1.5 weeks of training. We use the same NCPN architecture and hyperparameters across all datasets (except for dropout, which is set to 0.25 on CIFAR-10 and 0.00 on ImageNet 64x64 and CelebA 64x64). All NCPN models have 73M parameters.

\subsection{Additional Figures and Tables}

\begin{table}
  \caption{Results on CelebA 64x64. Negative log-likelihood (NLL) is in bits per dimension. Lower is better. *NLL with $\pi=0$ is equivalent to NLL of the original data.}
  \centering
  \label{table:celeba}
  \begin{tabular}{lccc}
    \toprule
    Model   & NLL   & NLL  & NLL  \\
            & $\pi=0$*    & $\pi=0.5$     & $\pi=1$   \\
    \toprule
    NCPN (ML)           & 2.25 & 3.72 & 4.35      \\
    NCPN (NCML VE)           & 2.22 & 3.63 & 4.21      \\
    NPCN (NCML sub-VP)           & 2.31 & 3.44 & 3.98      \\
    NCPN (NCML VP)           & 2.48 & 3.14 & 3.67      \\
    \bottomrule
  \end{tabular}
\end{table}

\begin{figure}
\centering
\includegraphics[width=\textwidth]{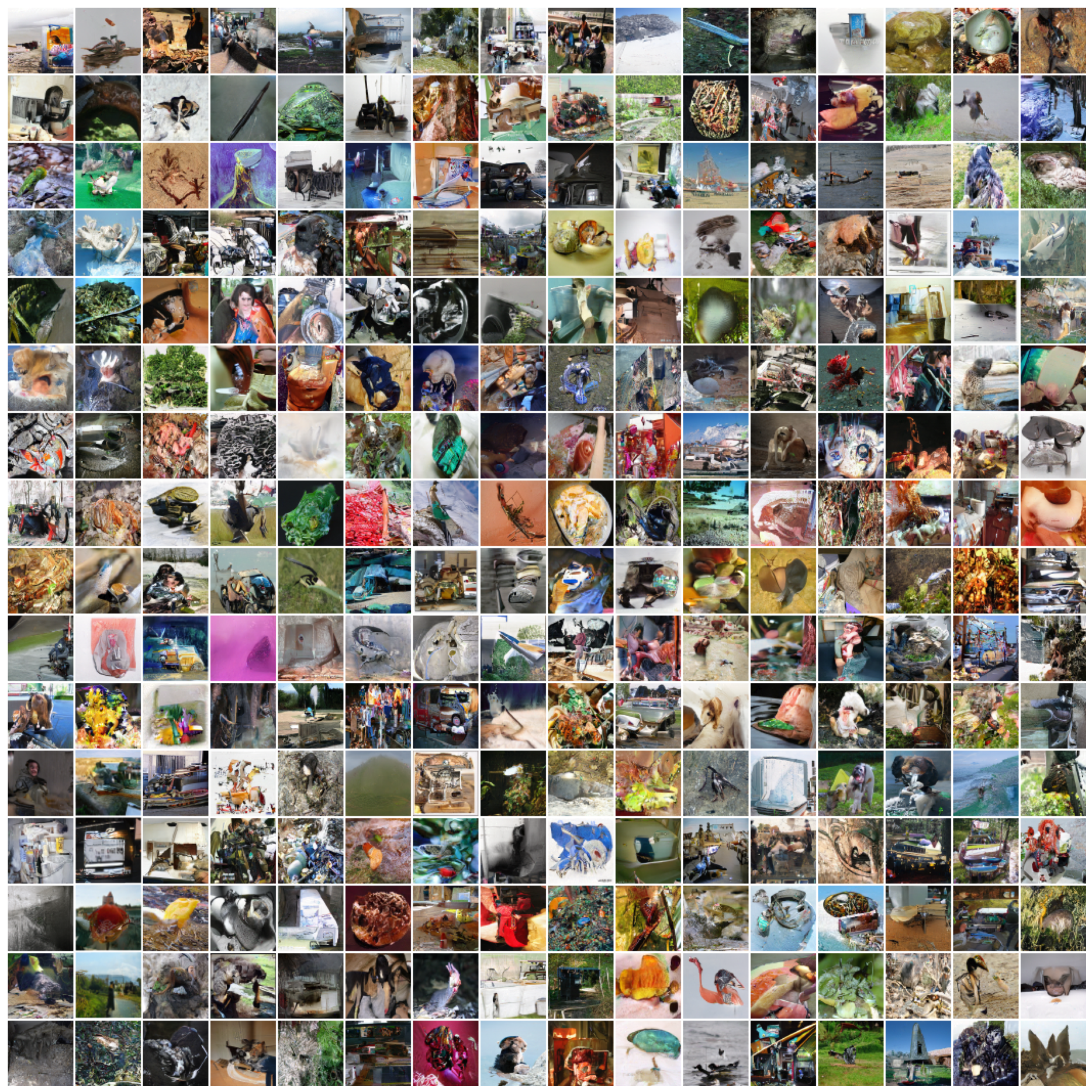}
\caption{Samples from NCPN trained on ImageNet 64x64, with $p_t$ as a variance preserving (VP) diffusion process.}
\label{fig:imagenet_vpsde}
\end{figure}

\begin{figure}
\centering
\includegraphics[width=\textwidth]{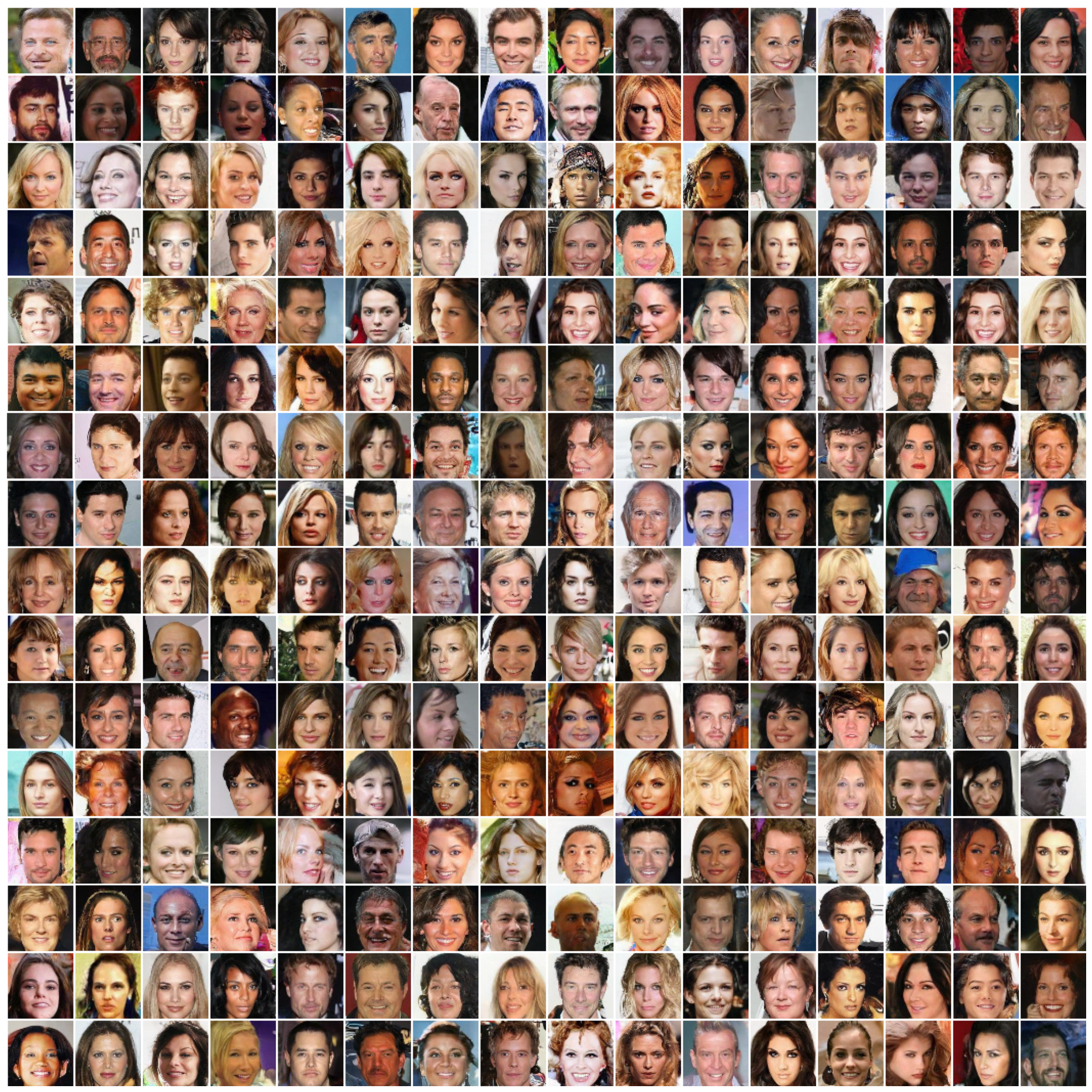}
\caption{Samples from NCPN trained on CelebA 64x64, with $p_t$ as a variance preserving (VP) diffusion process.}
\label{fig:celeba_vpsde}
\end{figure}

\begin{figure}
\includegraphics[width=\textwidth]{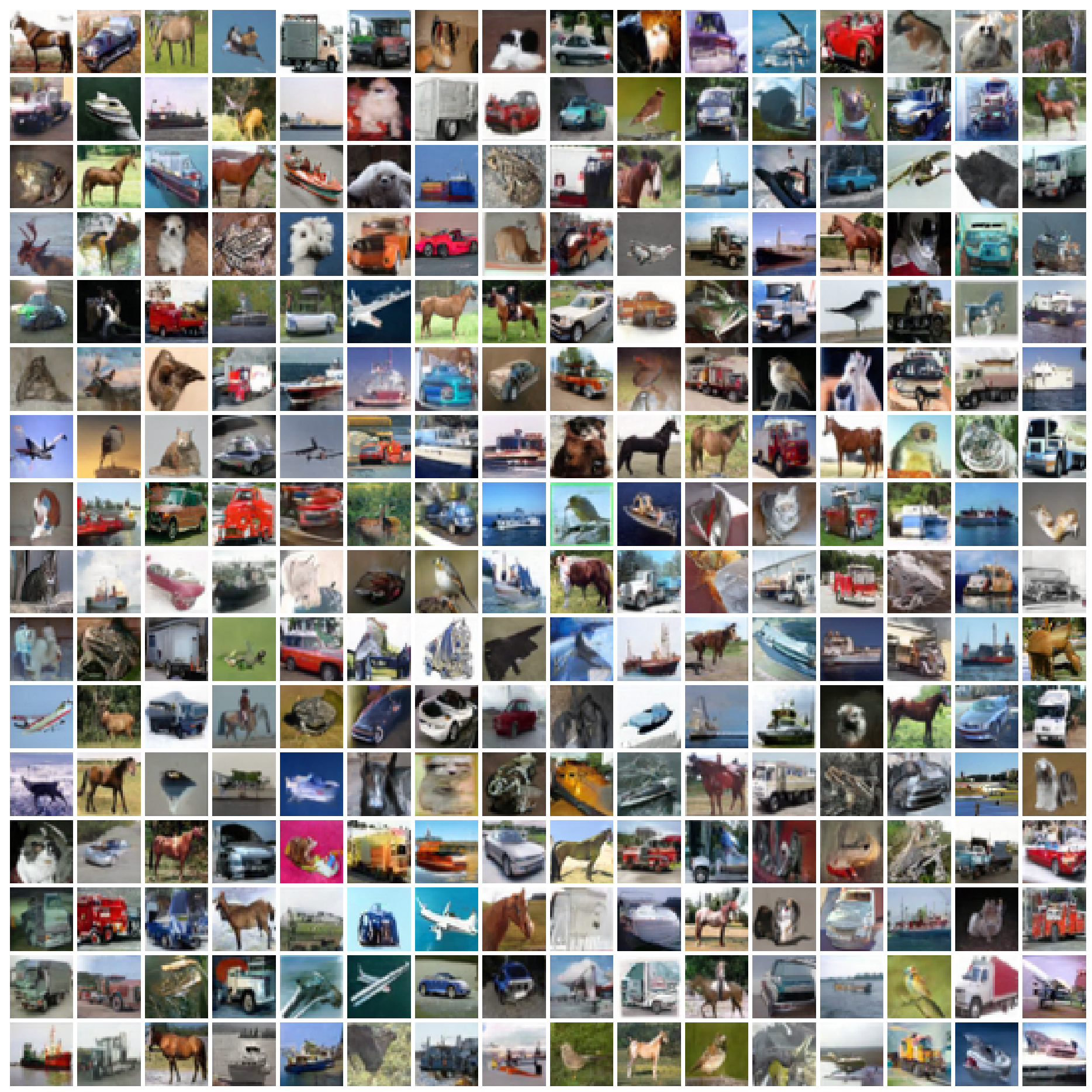}
\caption{Samples from NCPN trained on CIFAR-10, with $p_t$ as a variance preserving (VP) diffusion process.}
\centering
\end{figure}

\begin{figure}
\includegraphics[width=\textwidth]{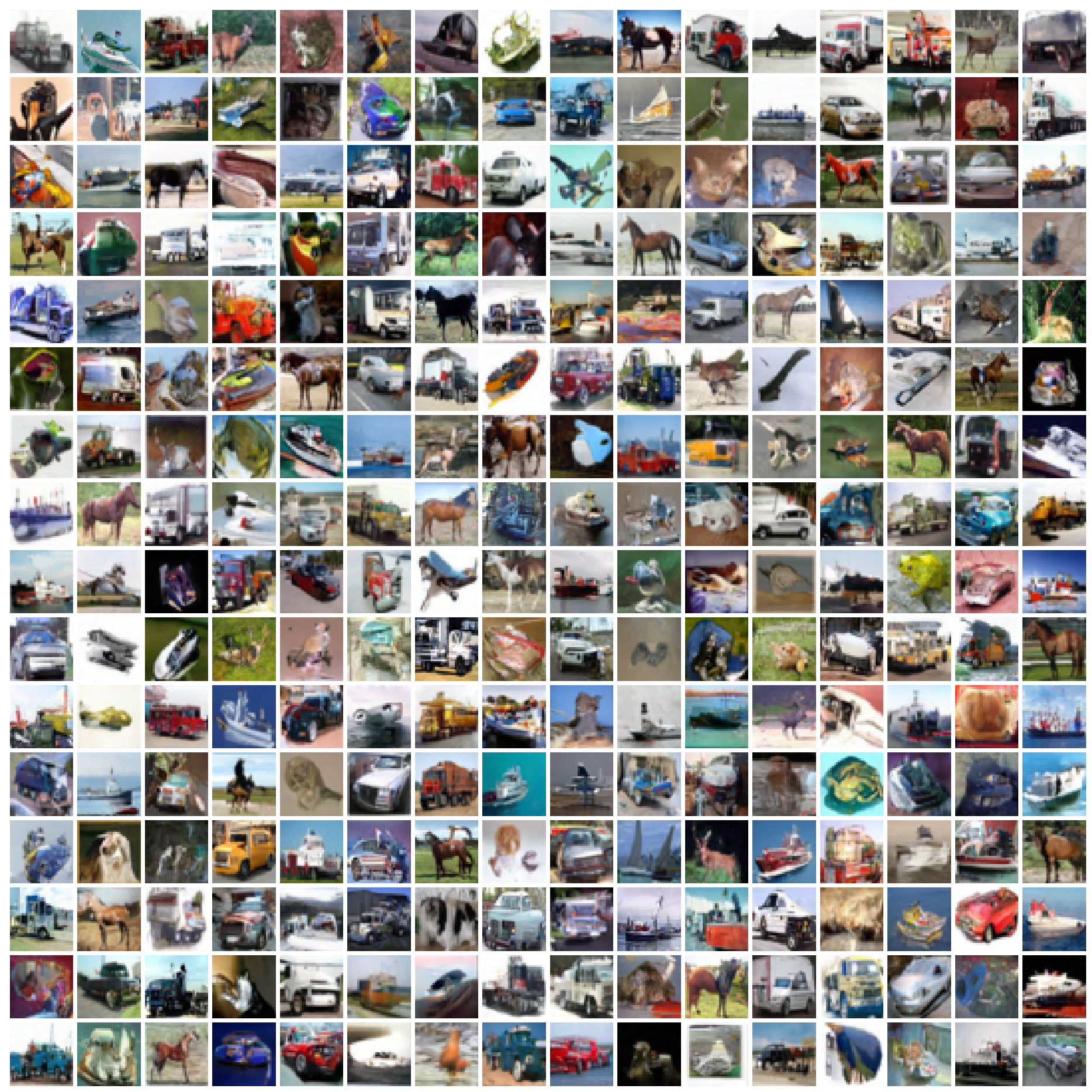}
\caption{Samples from NCPN trained on CIFAR-10, with $p_t$ as a sub-variance preserving (sub-VP) diffusion process.}
\centering
\end{figure}

\begin{figure}
\includegraphics[width=\textwidth]{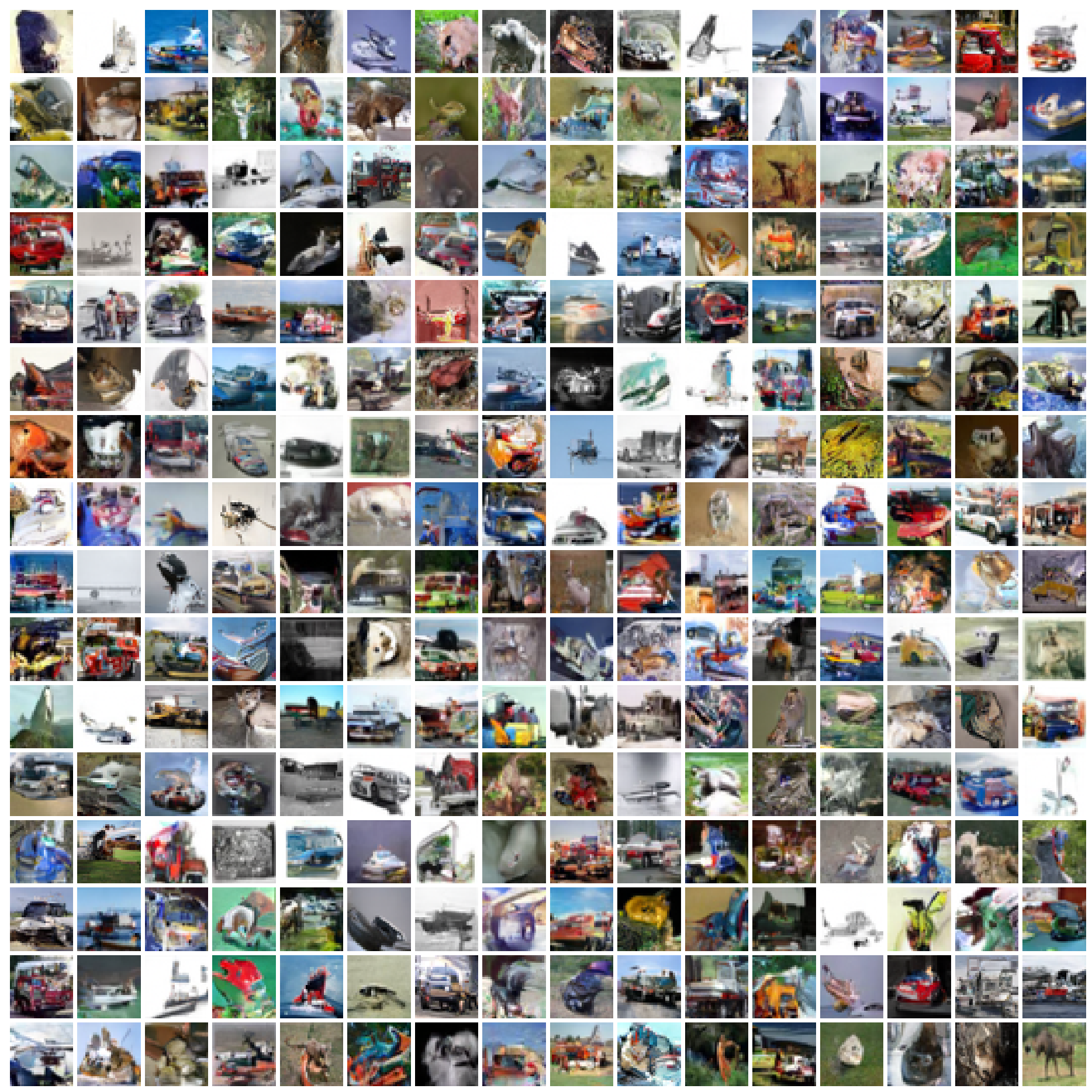}
\caption{Samples from NCPN trained on CIFAR-10, with $p_t$ as a variance exploding (VE) diffusion process.}
\centering
\end{figure}

\end{document}